\documentclass[conference]{IEEEtran}
\IEEEoverridecommandlockouts
\usepackage{cite}
\usepackage{amsmath,amssymb,amsfonts}
\usepackage{algorithmic}
\usepackage{graphicx}
\usepackage{textcomp}
\usepackage{xcolor}

\usepackage{booktabs}
\usepackage{algorithm}
\usepackage{algorithmic}
\usepackage{footnote}
\usepackage{multirow}
\usepackage{mathtools}
\usepackage{amssymb}
\usepackage{hyperref}
\hypersetup{
colorlinks=true,
linkcolor=blue,
citecolor=blue
}

\def\BibTeX{{\rm B\kern-.05em{\sc i\kern-.025em b}\kern-.08em
    T\kern-.1667em\lower.7ex\hbox{E}\kern-.125emX}}
\begin{document}

\title{HAIFIT: Human-to-AI Fashion Image Translation}

\author{\IEEEauthorblockN{Jianan Jiang}
\IEEEauthorblockA{\textit{Hunan University}\\
Changsha, Hunan, China \\
jiangjn22@hnu.edu.cn}
\and
\IEEEauthorblockN{Xinglin Li}
\IEEEauthorblockA{\textit{Hunan University}\\
Changsha, Hunan, China \\
lixinglin@hnu.edu.cn}
\and
\IEEEauthorblockN{Weiren Yu}
\IEEEauthorblockA{\textit{University of Warwick}\\
Coventry, CV4 7AL, UK \\
Weiren.Yu@warwick.ac.uk}
\and
\IEEEauthorblockN{Di Wu}
\IEEEauthorblockA{\textit{Hunan University}\\
Changsha, Hunan, China \\
dwu@hnu.edu.cn}
}

\maketitle

\begin{abstract}
In the realm of fashion design, sketches serve as the canvas for expressing an artist's distinctive drawing style and creative vision, capturing intricate details like stroke variations and texture nuances. The advent of sketch-to-image cross-modal translation technology has notably aided designers. However, existing methods often compromise these sketch details during image generation, resulting in images that deviate from the designer's intended concept. This limitation hampers the ability to offer designers a precise preview of the final output. To overcome this challenge, we introduce HAIFIT, a novel approach that transforms sketches into high-fidelity, lifelike clothing images by integrating multi-scale features and capturing extensive feature map dependencies from diverse perspectives. Through extensive qualitative and quantitative evaluations conducted on our self-collected dataset, our method demonstrates superior performance compared to existing methods in generating photorealistic clothing images. Our method excels in preserving the distinctive style and intricate details essential for fashion design applications. In addition, our method also has obvious advantages in model training and inference speed, contributing to reducing designers' time costs and improving design efficiency. Code is available at \url{https://github.com/ExponentiAI/HAIFIT/}.
\end{abstract}



\section{Introduction}
\label{sec:intro}

In the creative journey of design, sketches serve as the foundational tool for conceptualizing and outlining new pieces, demonstrating the unique styles of different designers. Designers initiate their process by crafting sketches that capture the basic structure of their envisioned creation. Subsequently, through iterative refinement, these sketches evolve, receiving modifications and intricate detailing to further refine the designs. Progressing from these refined stages, designers infuse life into their sketches by adding colors, aiming to impart depth and realism to their representations. Throughout this process, a robust sketch-to-image method plays a crucial role. It not only allows designers to preview renderings of colored sketches, aiding in refining the sketchwork, but also facilitates the generation of diverse image styles to inspire their designs.

\begin{figure}
\centering
\includegraphics[width=0.95\columnwidth]{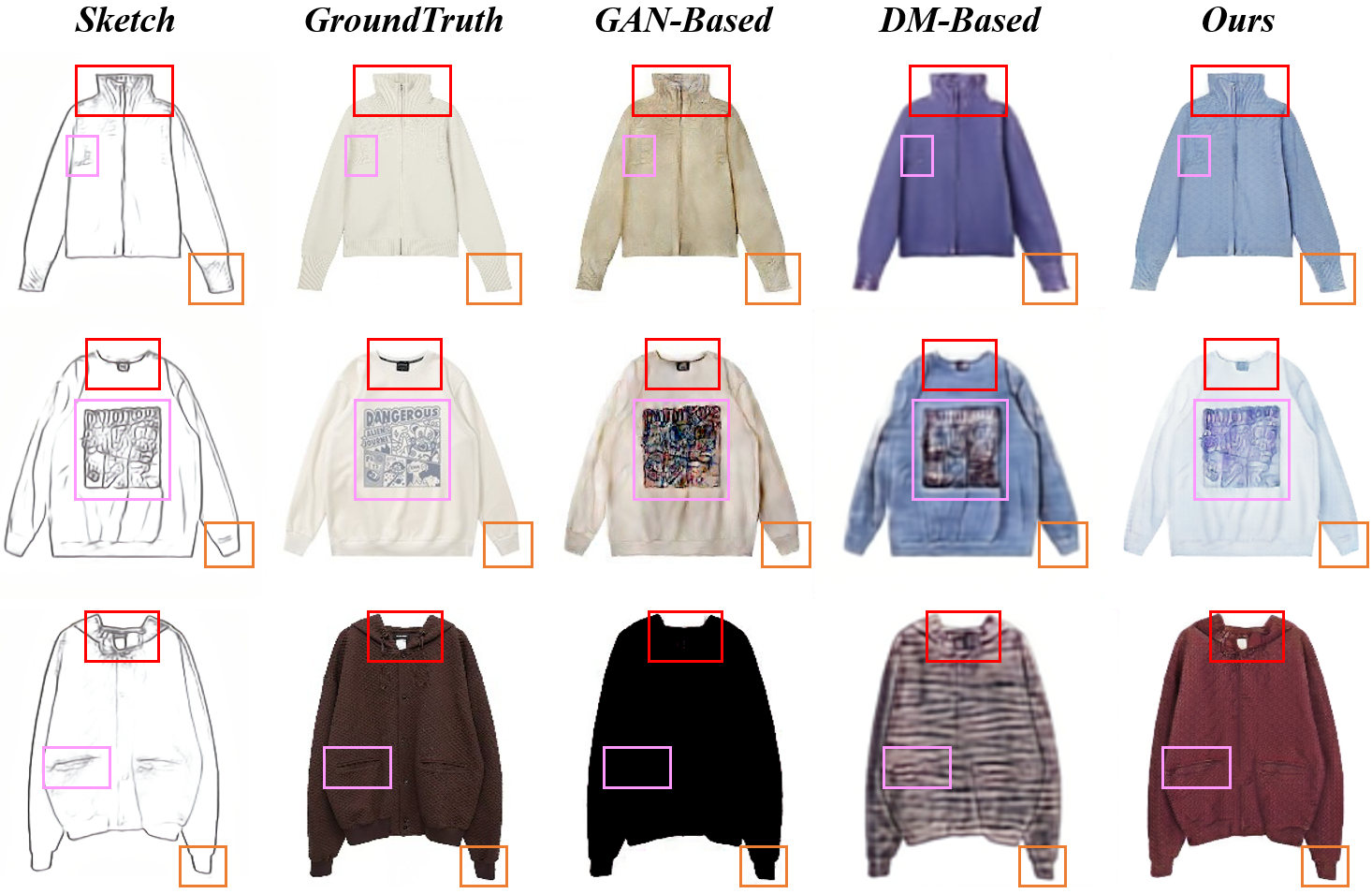}
\caption{The performance of different methodological approaches~\cite{wang2018high, li2023bbdm} compared to ours in handling sketch details (zooming in for better visual).}
\label{fig:back}
\end{figure}

In the earlier research, methods like Sketch2Photo~\cite{chen2009sketch2photo} synthesized realistic images by amalgamating object and background images derived from provided sketches. They laid the groundwork for deep learning techniques. The emergence of Generative Adversarial Networks (GAN)~\cite{goodfellow2014generative} marked a significant shift in this domain, with methods such as BicycleGAN~\cite{zhu2017toward}, pix2pix~\cite{isola2017image}, and others~\cite{huang2018multimodal, liu2020unsupervised, kim2020u-gat-it, wu2023styleme} gaining prominence. Although these methods have showcased notable results, preserving sketch details remains a challenge, as illustrated in Fig.~\ref{fig:back}. Through our comparison of two classic algorithms based on GAN~\cite{wang2018high} and Diffusion Model (DM)~\cite{li2023bbdm}, we discovered significant hurdles in retaining sketch details. Specific elements such as clothing necklines, pockets, sleeves, and other intricate details prove challenging to effectively express in the lines of the designer's sketches. This limitation significantly hampers designers' ability to preview and enhance finer details and refinements within sketches. 

In recent years, with the introduction of the diffusion model~\cite{ho2020denoising}, a novel and improved image generation algorithm has swiftly emerged. Some works, such as DDIM~\cite{song2020denoising}, employ skip sampling to expedite the model's inference process, while LDM~\cite{rombach2022high} utilizes feature mapping to execute the diffusion process in the latent space, effectively mitigating the issue of high resource requirements for model training. Additionally, Palette~\cite{saharia2022palette} and BBDM~\cite{li2023bbdm} showcase the efficacy of diffusion models in image translation tasks. However, these methods heavily rely on extensive data for learning and demand substantial computing resources. Consequently, they still face significant challenges in model training, inference, and fine-tuning, resulting in high time and resource costs, which may not be user-friendly for designers.

To address these challenges, we introduce HAIFIT (\textbf{H}uman-to-\textbf{AI} \textbf{F}ashion \textbf{I}mage \textbf{T}ranslation), a GAN-based creative generation network designed for sketch-to-image translation in fashion design. HAIFIT aims to empower designers to foster inspiration by preserving as much detail as possible from their sketches in the generated images. Specifically, HAIFIT focuses on converting clothing sketches into high-quality realistic clothing images using a pyramidal generative adversarial network. To facilitate this conversion, we have developed a customized \textit{Multi-scale Feature Fusion Encoder} (MFFE) that maps the reference hand-drawn sketches and generated images into a shared latent space. MFFE comprises a \textit{Shallow Convolution Module} (SCM) and an \textit{Abstract Feature Representation Module} (AFRM), which enables the capture of long-term dependencies between different regions and improves the global correlation in feature representations. Furthermore, we enhance the fine-grained features among generated images of various scales by integrating the Cross-level Skip Connection Module (CSCM) within the feature space across different layers. By employing multi-scale discriminators, we ensure the preservation of global structural integrity while emphasizing fine details, resulting in more realistic generated images. 

The key \textbf{contributions} of our work are summarised as follows:
\begin{itemize}
\item \textbf{Methodology.} To the best of our knowledge, we are the pioneers in treating sketch features as sequence inputs, ensuring consistency by inputting features from both forward and reverse feature sequences. This innovative approach enhances the representation of details in the generated images. Moreover, we are the first to propose a Pyramid Diffusion Model, which uses a progressive generation approach to improve the quality of model generation.
\item \textbf{Structure.} We introduce the LSTM-based MFFE to enhance the representation of global features and ensure coherence and consistency in the latent space. Additionally, our CSCM strategically directs the model's attention to finer sketch details, contributing to improved and more nuanced sketch feature representation.
\item \textbf{Datasets.} We construct a background-free dataset comprising fashion clothes and their corresponding sketches for cross-modal image translation tasks, filling the gap in this field.
\item \textbf{Performance.} Our method demonstrates superior performance in extensive quantitative and qualitative experiments compared to various baselines.
\end{itemize}

The rest of the article is structured as follows: Sec.~\ref{sec:work} provides an overview of the work and technological progress related to image translation. Sec.~\ref{sec:s2i} presents the technical details of our approach. Sec.~\ref{sec:diff} introduces a version of our method extended to a diffusion model. Sec.~\ref{sec:experiments} presents the experimental results of our approach. Sec.~\ref{sec:conclusion} summarizes our work.

\begin{figure*}[t]
\centering
\includegraphics[width=\linewidth]{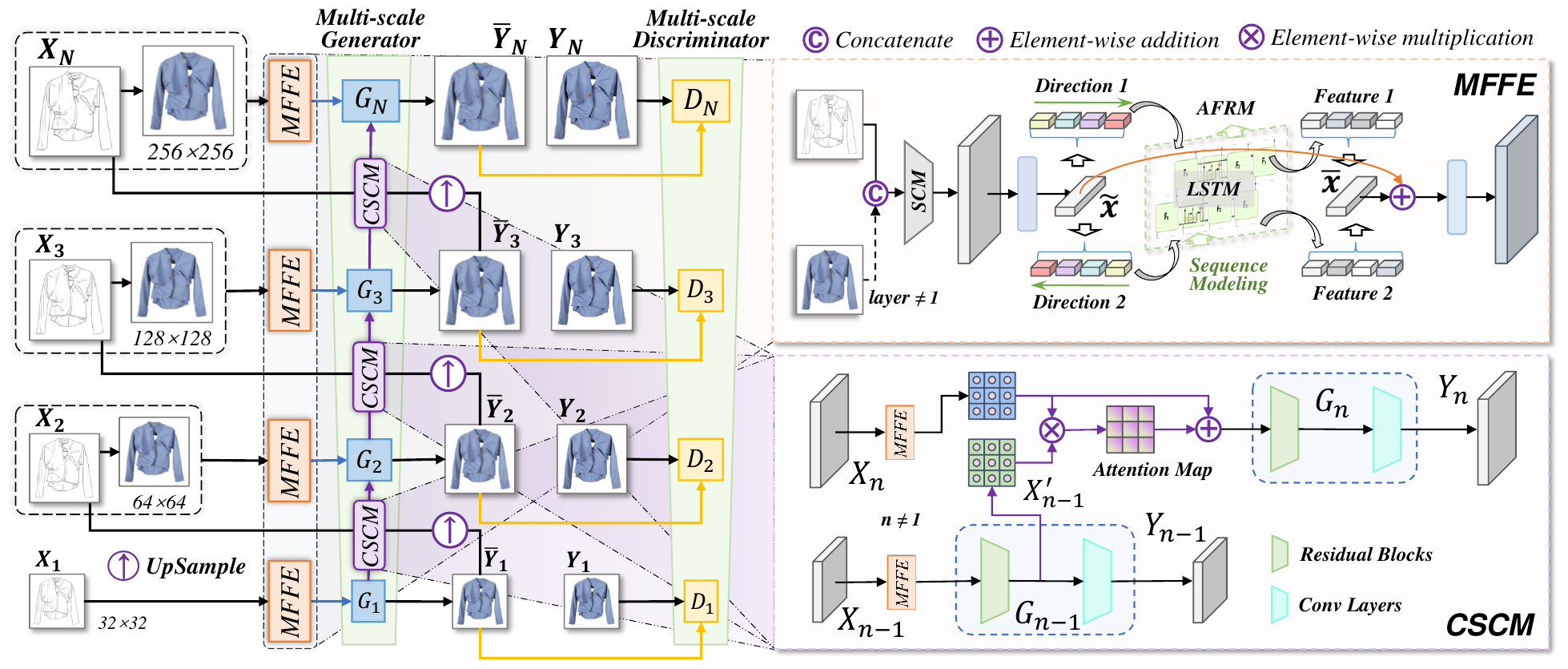}
\caption{\textbf{(i)} [Left] The illustration of our HAIFIT architecture. Given $M$ paired sketch-image samples $\{(x_i,y_i) \in (X,Y)\}^{M}_{i=1}$, our goal is to map images from hand-drawn sketches domain $X$ to realistic images domain $Y$, while preserving as many line details from the sketch $X$ as possible in the generated image $\overline{Y}$. \textbf{(ii)} [Top Right] An overview of the Multi-scale Feature Fusion Encoder (MFFE). \textbf{(iii)} [Bottom Right] An overview of the Cross-level Skip Connection Module (CSCM).}
\label{fig:s2i}
\end{figure*}


\section{Related Work}
\label{sec:work}

\subsection{Sketch-to-Image Generation}
The task of sketch-to-image generation falls within the realm of image translation, a pivotal objective in computer graphics and computer-human interaction. In the domain of fashion clothing design, having an effective algorithm that can accurately generate diverse clothing images based on designers' sketches is invaluable. Such algorithms enable designers to express their creative ideas effectively, save development time, and enhance design efficiency. Early methods, like Photosketcher~\cite{eitz2011photosketcher}, utilized local features for image retrieval and employed graphics techniques to segment and stitch images together to achieve sketch-to-image generation. Subsequently, with the development of convolutional neural networks~\cite{krizhevsky2012imagenet, szegedy2015going, simonyan2014very, he2016deep}, deep learning-based methods have become the mainstream approach for achieving sketch-to-image generation. Isola et al.~\cite{isola2017image} introduced a unified framework for image-to-image translation models adaptable to various scenarios. This framework facilitates mapping segmented images to natural images or converting edge maps to color images, among other applications. Building upon this, Pix2piHD~\cite{wang2018high} utilized multi-scale generators and discriminators to achieve high-resolution image translation. Several other methods~\cite{huang2018multimodal, alami2018unsupervised, liu2020unsupervised, kim2020u-gat-it} have improved image generation performance by incorporating attention networks, innovative adversarial loss functions, or different image features. However, these methods still encounter challenges in handling complex strokes and refining synthetic structures to produce realistic images. In summary, synthesizing an image that faithfully represents the content of a sketch while maintaining consistent styles is paramount.

\subsection{Generative Adversarial Networks}
The Generative Adversarial Network (GAN) comprises two components: the generator and the discriminator. It leverages adversarial training to enhance the generator's ability to generate realistic images, yielding significant achievements in image generation, restoration, and various other domains. Initially proposed by Goodfellow et al.~\cite{goodfellow2014generative}. For generating realistic data, GANs have since undergone several extensions and improvements. Mirza et al.~\cite{mirza2014conditional} extended the GAN to enable the generation of images with specific conditions, allowing users to control attributes such as image categories. Radford et al.~\cite{radford2015unsupervised} introduced convolutional neural networks into both the generator and discriminator, thereby enhancing GAN's ability to capture spatial image features. Zhu et al.~\cite{zhu2017unpaired} achieved cross-domain image conversion by introducing a cycle consistency loss function, enabling realistic transformations from one domain to another, such as from horse to zebra. Karras et al.~\cite{karras2019style} achieved stable control over the style and diversity of generated images by introducing style variables. Brock et al.~\cite{brock2018large} further scaled up the GAN and improved training methods to generate high-resolution images. Additionally, methods like WGAN~\cite{arjovsky2017wasserstein}, PGGAN~\cite{karras2017progressive}, and SAGAN~\cite{zhang2019self} have introduced new optimization losses, multi-layer resolution training strategies, and self-attention mechanisms to achieve more stable GAN training and further enhance the quality and diversity of generated images. In general, the continuous development and evolution of GANs have unlocked significant potential in the field of image generation, transforming the generation of realistic images from a concept into reality. Compared to the recent diffusion models~\cite{saharia2022palette, li2023bbdm}, GANs possess the advantages of low hardware costs and fast synthesis speed, thus continuing to play a key role in the design field.

\subsection{Diffusion Models}
Different from GANs, the diffusion model~\cite{ho2020denoising} realizes image generation from a novel perspective. It comprises two stages: the forward noise-adding stage and the reverse denoising stage. In the forward noise-adding stage, Gaussian noise is continuously added to the initial input data to make it approximately Gaussian. In the reverse denoising stage, the model is tasked with recovering the original input data by learning to gradually reverse the denoising process. Diffusion models are widely appreciated for the quality and diversity of the samples they generate, but they also impose a severe computational burden and time cost. DDIM~\cite{song2020denoising} addresses the issue of single-step sampling in the inference process through skip sampling. Subsequent advancements like ADM~\cite{dhariwal2021diffusion} further demonstrate the superiority of the diffusion model over traditional GAN methods in image synthesis. Another notable approach by Ho et al.~\cite{ho2022classifier} proposes a diffusion model training method without classifier guidance. Palette~\cite{saharia2022palette} introduces a unified framework for image-to-image translation based on the conditional diffusion model, achieving superior results across multiple image generation subtasks. However, they still encounter significant model training costs. LDM~\cite{rombach2022high} addresses this by significantly reducing model training costs and opening up new avenues for subsequent research by placing the diffusion process in latent space. DiT~\cite{peebles2023scalable} explores a new class of diffusion models based on transformer architecture, enhancing the scalability and quality of diffusion models. Overall, these studies continuously advance the field of image generation and offer new possibilities for natural and realistic image synthesis.


\section{Methodology}
\label{sec:s2i}

In this study, we present HAIFIT, as depicted in Fig.~\ref{fig:s2i}, which draws on the training strategy of PGGAN~\cite{karras2017progressive} and comprises two core components: the \textit{Multi-scale Feature Fusion Encoder} (MFFE), and the \textit{Cross-level Skip Connection Module} (CSCM). The MFFE is used to get global contour features and abstract intent features. Then the feature maps are resized and concatenated at multiple scales before being fed into the multi-scale pyramid generator, which can produce realistic images with complete structures and exquisite details.

\subsection{Multi-scale Feature Fusion Encoder}
The \textit{Multi-scale Feature Fusion Encoder} (MFFE) consists of two modules: the \textit{Shallow Convolutional Module} (SCM) and the \textit{Abstract Feature Representation Module} (AFRM). As illustrated in Fig.~\ref{fig:s2i} (ii), the SCM employs a convolutional network to learn essential global contour information, such as the structure of the object. Simultaneously, inspired by the LSTM architecture ~\cite{hochreiter1997long}, we design the AFRM to learn long-range dependencies in key areas of the input sketch from various perspectives, which can effectively capture style features and geometric features of strokes. By combining these features, we can effectively capture the consistency of objects' complete structure and intricate textures within the latent space.

As shown in Fig.~\ref{fig:s2i} (i), in the bottom layer, only the sketch features are inputted. In the other layers, we perform element-wise addition between the generated image features and the sketch features, and then input the resulting summed features into the MFFE. In the SCM we employ a sequence of convolutional layers, denoted as $Con$, to extract the global contour features and shape information $\widetilde{x}$ from the input sketch $x_s$. Here, $x_s \in \mathcal{R}^{b \times 3 \times m \times m}$, $\widetilde{x} \in \mathcal{R}^{b \times 256 \times m/4 \times m/4}$, $b$ is the batch size, and $m$ depends on the network layer, so we can express this process as $\widetilde{x}=Con(x_{s})$. Specifically, the SCM extracts features in a blockwise manner, which will disrupt the continuity of strokes and result in pixel-level contour features. Since sketches are continuous and composed of multiple lines in different directions, it is crucial to consider the correlation between adjacent lines in order to effectively learn the line features of complex sketches.

To address this issue, we introduce the AFRM to learn the feature $\widetilde{x}$. The AFRM consists of a two-layer LSTM network, capturing abstract drawing intent features from four directions. After downsampling $\widetilde{x}$, we obtain the feature $\widetilde{x'} \in \mathcal{R}^{b \times 256 \times 4 \times 4}$. We then swap feature positions after merging its second and third dimension features to obtain $\mathcal{R}^{4 \times b \times 1 \times 1024}$ and divide it into four parts, namely $\widetilde{x}_{0}$, $\widetilde{x}_{1}$, $\widetilde{x}_{2}$, and $\widetilde{x}_{3}$, with dimensions $\mathcal{R}^{b \times 1 \times 1024}$. These features are then individually sequence inputted into the LSTM network, generating features $\overline{x}_{0}$, $\overline{x}_{1}$, $\overline{x}_{2}$, and $\overline{x}_{3}$ by recursively processing the input features from different directions. Next, we concatenate the features $\overline{x}_{0}$, $\overline{x}_{1}$, $\overline{x}_{2}$, and $\overline{x}_{3}$ to obtain the feature $\overline{x'} \in \mathcal{R}^{4 \times b \times 1 \times 1024}$. Afterward, we restore the dimension of $\overline{x'}$ to the dimension of $\widetilde{x'}$. Finally, we apply a 1$\times$1 convolutional layer along the channel dimension to obtain the output feature $\overline{x} \in \mathcal{R}^{b \times 256 \times 4 \times 4}$.

\begin{figure}[t]
\centering
\includegraphics[width=\columnwidth]{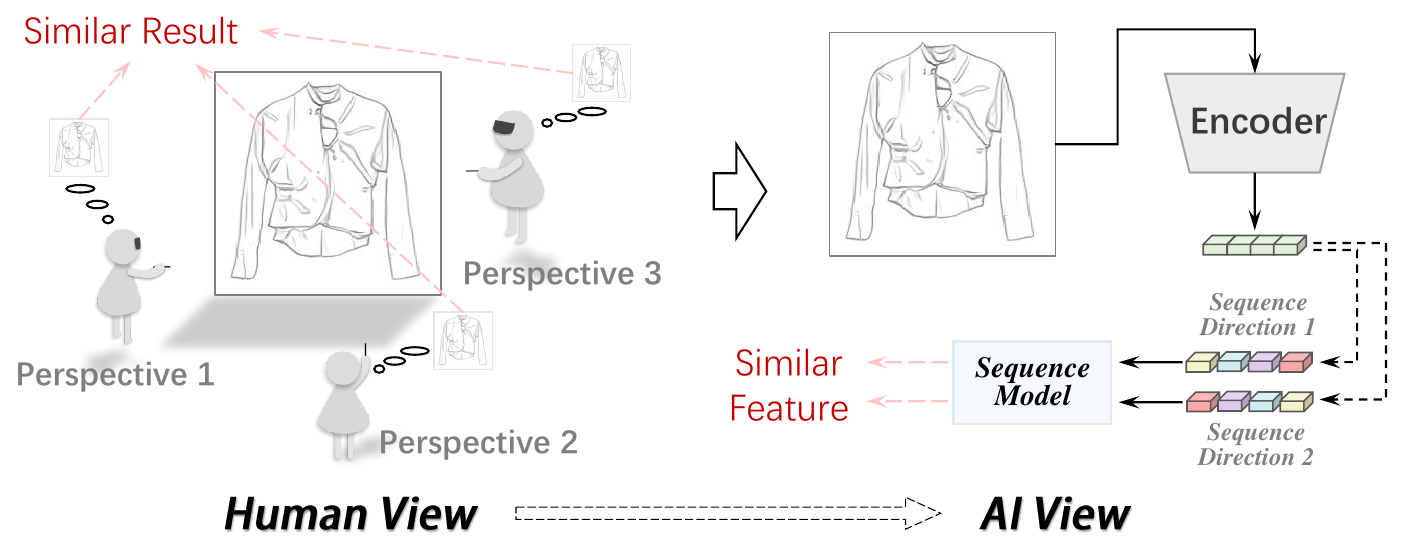}
\caption{From the Human View to the AI View, different perspectives yield consistent results.}
\label{fig:view}
\end{figure}

The MFFE is employed to learn features from different perspectives of the sketches, which enables us to capture variations in lines and abstract painting intentions in crucial areas of the sketches. In the task of sketch-to-image generation, it is vital for the encoder to acquire diverse line patterns and painterly intent in clothing patterns and details. Failure to do so can severely impact the visual quality of the synthesized image, even if the overall generation quality is satisfactory. As a result, to further ensure the consistency of the generated images, we introduce the concept of reverse sketch feature input. This is because a sketch should yield consistent results regardless of the direction it is interpreted from, which can be referred to Fig.~\ref{fig:view}. We also provide the structural details related to the AFRM in MFFE in Fig.~\ref{fig:s2i} (ii). We serialize the feature $\widetilde{x}$ obtained from the SCM and input it into the AFRM through forward and reverse feature inputs. The output features $\overline{x}_{for}^{\prime}$ and $\overline{x}_{rev}^{\prime}$ are then fused to obtain the unified sketch abstract stroke feature $\overline{x}$, which is subsequently used in the later stages of the model. Therefore, the aforementioned process can be summarized as follows:
\begin{equation}
\label{eq:mffe}
\begin{split}
\overline{x}_{for}^{\prime} &= Cat(LSTM[\widetilde{x}_{0},\widetilde{x}_{1},\widetilde{x}_{2},\widetilde{x}_{3}]) \\
\overline{x}_{rev}^{\prime} &= Cat(LSTM[\widetilde{x}_{3},\widetilde{x}_{2},\widetilde{x}_{1},\widetilde{x}_{0}]) \\
\overline{x} &= Conv_{1\times1}(Cat(\overline{x}_{for}^{\prime}, \overline{x}_{rev}^{\prime}))
\end{split}
\end{equation}

Finally, we fuse the global structure feature $\widetilde{x}$ and abstract intention feature $\overline{x}$ through a differentiable parameter $\zeta$ with the initial value of 1.0 to get the embedded sketch feature $x$:
\begin{equation}
\label{eq:x}
x=\zeta\widetilde{x}+\overline{x}
\end{equation}

\subsection{Cross-level Skip Connection Module}
In order to effectively capture intricate details in the key areas of the sketch, we employ a pyramid generative learning approach~\cite{karras2017progressive} with a \textit{Cross-level Skip Connection Module} (CSCM). Our model comprises a pyramid of generators $G_n \in \{G_1, G_2, \dots, G_N\}$, which are trained using a set of multi-scale images $x_i$ represented as ${x^1_i, x^2_i, \dots, x^N_i}$. Each generator $G_n$ is responsible for generating a realistic image $\overline{y}^n_i$. By adversarial training, the generator $G_n$ aims to deceive the associated discriminator $D_n \in \{D_1, D_2, \dots, D_N\}$, and the $D_n$ is trained to distinguish between the generated realistic images and real images. The generation of a realistic image sample starts at the coarsest scale $64\times64$ and sequentially passes through all generators, progressing toward the finest scale $256\times256$. At the coarsest scale, the MFFE takes a minibatch of the sketch feature maps $X_{n}$ as input. The effective receptive field at this level is typically half the height of the finer image. Each of the generators $G_n$ at finer scales adds details that are not generated by the previous scales. Therefore, in addition to the feature maps $X_{n}$, each multi-scale feature fusion encoder accepts an upsampled version of the generated image features from the coarser scale. Here, $X_{n} \in \mathcal{R}^{b \times 3 \times m \times m}$, $Y_{n-1} \in \mathcal{R}^{b \times 3 \times m/2 \times m/2}$. This pyramid structure allows us to capture features at multiple scales and better handle the complex details of the sketch.
\begin{equation}
\label{eq:l_style}
Y_{n} =
\begin{cases} 
G_n(X_{n}),  & \mbox{if \textit{n} = 1} \\
G_n(X_{n},Y_{n-1}), & \mbox{if \textit{n} \textgreater 1}
\end{cases}
\end{equation}

The generator $G_n$ consists of a residual block and multiple convolutional layers. In the coarsest layer, the residual block comprises four convolutional layers. In addition, in the pyramid network, as the number of pixels increases from bottom to top, the residual block progressively adds one convolutional layer at each level. This progressive increment facilitates the learning of sketch feature representations. Through numerous experiments, it has been observed that incorporating the CSCM between different layers in the pyramid network can enhance the quality of the generated images, which is illustrated in Fig.~\ref{fig:s2i} (iii). In detail, We extract the output feature ${X'}_{n-1}$ from the last layer of the residual block in generator $G_{n-1}$ and concatenate it with the output feature ${X}_{n}$ from the $n$-th layer of the MFFE. Subsequently, we use $\overline{X}_{n}$ as the input for generator $G_{n}$, where $\overline{X}_{n} \in \mathcal{R}^{b \times 256 \times m\times m}$. In principle, the multi-level features of the same synthetic image should exhibit similarity. By employing cross-level feature operations, we can extract similar details from the features generated by different level generators and incorporate them as attention maps into the new feature map $X_n$. This facilitates image translation, resulting in more detailed and high-quality images.
\begin{equation}
\label{eq:Ygen}
\begin{split}
\overline{X}_{n} &= X_{n} + X_{n} \cdot X'_{n-1} \\
Y_{n} &= G_n(\overline{X}_{n},Y_{n-1}), n=2, 3, 4
\end{split}
\end{equation}

\begin{algorithm}[t]
\caption{The Diffusion HAIFIT Training Process.}
\begin{algorithmic}[1]
\label{alg_diff}
\REQUIRE $S$: Input Sketches
\REQUIRE $E$: Sketch Encoder network
\REQUIRE $\epsilon$: Gaussian Noise
\REQUIRE $step$: Model training steps
\REQUIRE $t\in [0, 1000]$: Diffusion moment
\REQUIRE $G_i \in \{G_1, G_2, G_3, G_4\}$: Generator networks
\REQUIRE $\bar{Y}_i \in \{\bar{Y}_1, \bar{Y}_2, \bar{Y}_3, \bar{Y}_4\}$: Generated Images

\STATE Randomly initialize $\epsilon$ as Input
\FOR{$i = 1$ to 4}
    \STATE Utilize $E$ to encode $S$ as supervised conditions
    \FOR{0 to $step$}
        \STATE Randomly initialize $t$
        \STATE Train $G_i$ to predict the $t$-th moment noise $\bar{\epsilon}_t$
        \STATE Compute loss between $\bar{\epsilon}_t$ and real noise $\epsilon_t$
        \STATE Update parameters of $G_i$ using gradient descent
    \ENDFOR
    \IF{$i < 4$}
        \STATE Freeze parameters in $G_i$ and use it for inference
        \STATE Upsample $\bar{Y}_i$ to double its original resolution and concatenate with new resolution $\epsilon$ as Input
    \ENDIF
\ENDFOR
\STATE Utilize all $\{G_i\}_{i=1}^4$ for inference generation
\end{algorithmic}
\end{algorithm}

\subsection{Loss Function}
Our full loss function consists of L1 loss, adversarial loss, style loss, and perceptual loss. We empirically set the following parameters: $\lambda_{l1}$=1.5, $\lambda_{adv}$=10.0, $\lambda_{per}$=0.1, $\lambda_{style}$=250.0, to keep stable training and higher-quality results.
\begin{equation}
\label{eq:losssi}
\mathcal{L_S} = \lambda_{l1}\mathcal{L}_{l1}+\lambda_{adv}\mathcal{L}_{adv}+\lambda_{style}\mathcal{L}_{style}+\lambda_{per}\mathcal{L}_{per}
\end{equation}

\noindent \textbf{L1 Loss.} 
We directly employ the $L1$-distance loss to enhance the similarity between the pixel features of the generated image $G(x)$ and the real image $y$.
\begin{equation}
\label{eq:l1}
\mathcal{L}_{l1}=\|y-G(x)\|_{1}
\end{equation}

\noindent \textbf{Adversarial Loss.}
To improve the quality of generated samples, we utilize the vanilla GAN~\cite{goodfellow2014generative} min-max loss for the generator $G$ and discriminator $D$ training process.
\begin{equation}
\label{eq:adv}
\mathcal{L}_{adv}=\mathbb{E}_y[\log D_Y(y)]+\mathbb{E}_x[\log(1-D_Y(G(x))]
\end{equation}

\noindent \textbf{Style Loss.}
We incorporate the $Gram$ matrices-based loss to better reproduce the texture details of input sketches $x$.
\begin{equation}
\label{eq:lstyle}
\begin{split}
\mathcal{L}_{style} &= \|Gram(y)-Gram(G(x))\|_{1}
\end{split}
\end{equation}

\noindent \textbf{Perceptual Loss.}
To ensure both semantic similarity between the generated image and its ground truth, as well as encourage the network to generate diverse images, we employ the pre-trained VGG-16~\cite{simonyan2014very} $\phi$ to extract $L$-levels features.
\begin{equation}
\label{eq:per}
\mathcal{L}_{per}=\sum_{i=1}^{L=5}\frac{1}{N_{i}}\| \phi_{i}(y)-\phi_{i}(G(x))\|_{1}
\end{equation}

\section{Pyramid Diffusion Model}
\label{sec:diff}

In recent years, the Diffusion models~\cite{ho2020denoising,song2020denoising,rombach2022high} have garnered significant attention from researchers due to their potent generation capabilities, leading to rapid advancements and widespread adoption across various fields. Building upon this, we extend our GAN-based HAIFIT model to incorporate a diffusion model version: the Pyramid Diffusion Model.

\begin{figure}[t]
\centering
\includegraphics[width=0.98\columnwidth]{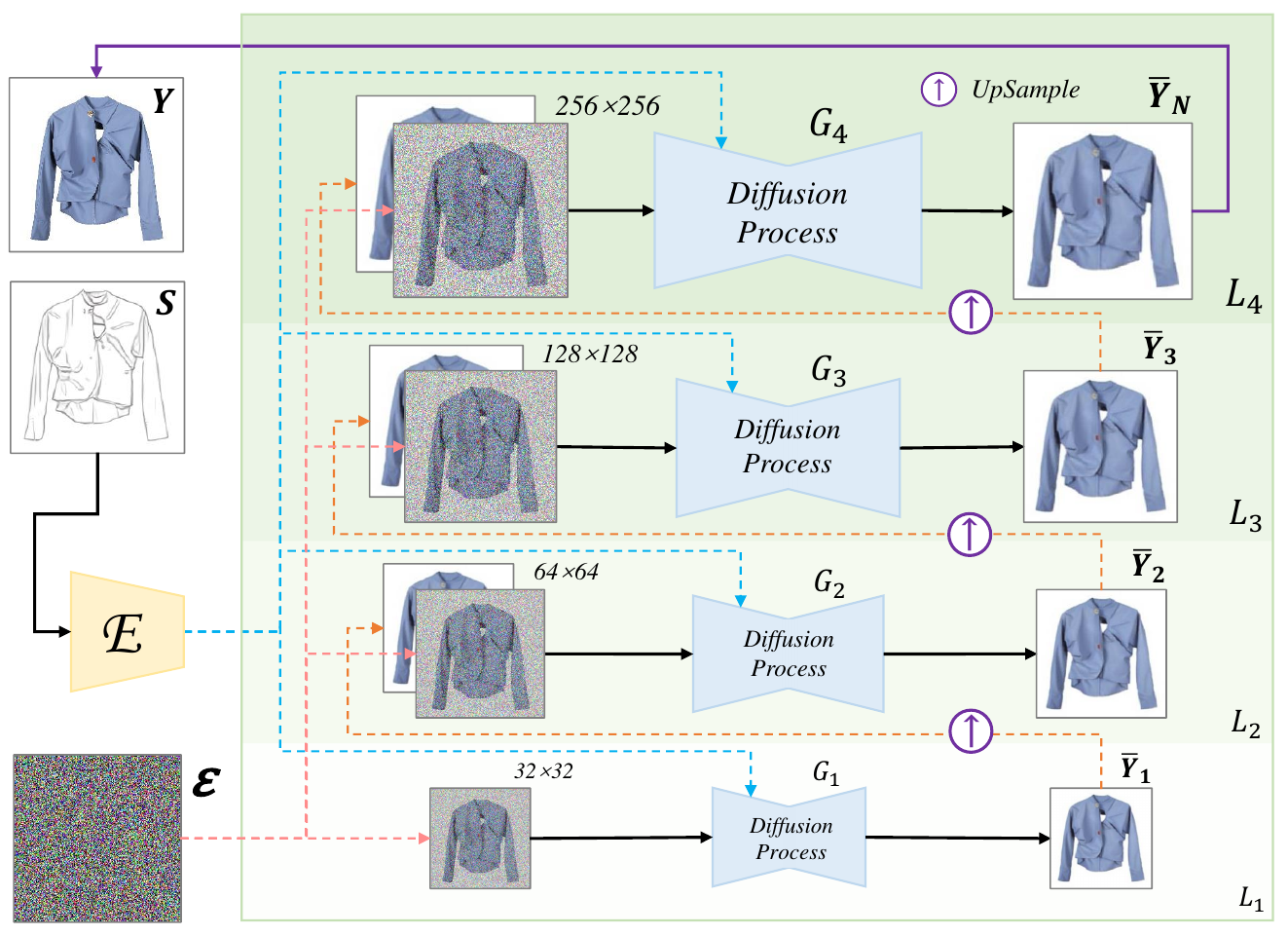}
\caption{The illustration of our HAIFIT with Diffusion model architecture.}
\label{fig:diff}
\end{figure}

\subsection{Preliminaries of DDPM}
The Denoising Diffusion Probabilistic Model (DDPM) comprises a forward process and a reverse process. In the forward process, noise $\epsilon$ is iteratively added to the original image $X_0$, resulting in a noisy image $X_T$. The noise $\epsilon$ added at each time step $t$ is sampled from a Gaussian distribution $\mathcal{N}(0, I)$. When $T$ is sufficiently large, we believe that $X_T$ approximately conforms to the Gaussian distribution, here we set $T=1000$. This transmission can be mathematically represented as $q(X_t|X_{t-1}) = \mathcal{N}(\sqrt{1-\beta_t}X_{t-1}, \beta_tI)$, $\{\beta_t\}_{t=0}^T \in (0, 1)$. Therefore, the noised image $X_t$ at $t$-th moment in the forward process can be expressed as:
\begin{equation}
X_t = \sqrt{\bar{\alpha}_t}X_0 + \sqrt{(1 - \bar{\alpha}_t)}\epsilon, \epsilon \sim \mathcal{N}(0, I)
\end{equation}

\noindent where $\alpha_t = 1- \beta_t$ and $\bar{\alpha}_t = \prod_{i=1}^t \alpha_i$.

In the reverse process stage, the noisy image $X_T$ is continuously denoised to revert it back to the original input image $X_0$. This is achieved by training a model $\theta$ to predict the noise $\epsilon$ added to at each time step $t$, thereby removing the corresponding noise $\epsilon$. Therefore, the approximate probability distribution of the image $X_{t-1}$ at time $t$ can be expressed as:
\begin{equation}
p_{\theta}(X_{t-1}|X_t) = \mathcal{N}(X_{t-1}; \mu_\theta(X_t, t), \sigma^2I)
\end{equation}

To summarize, the optimization objective function of the diffusion model training process is:
\begin{equation}
\mathcal{L}(\theta) = \mathbb{E}_{x,\epsilon\sim\mathcal{N}(0,1),t}\Big[\|\epsilon-\epsilon_\theta(x_t,t)\|_2^2\Big]
\end{equation}

\subsection{Diffusion HAIFIT}
As presented in Fig.~\ref{fig:diff}, We utilize the diffusion method to hierarchically train each layer in HAIFIT. Beginning from the $L_1$ layer, we progressively train upwards to the $L_4$ layer. Upon completing the training of the previous layer network, we freeze the parameters of the network and add a new layer network upwards for further training until all layers of the model are trained. Specifically, we initialize random noise $\epsilon$ and then train $G_1$ to predict the noise $\epsilon_t$ at each moment $t$ and perform denoising operations on $X_t$ until $G_1$ effectively simulates the noise value added at each moment $t$, where $t \in [0, 1000]$. During this process, we employ an encoder $E$ to encode the features of the corresponding sketch $S$ for supervised diffusion model training. This encoder is a backbone based on a pre-trained ResNet~\cite{he2016deep}. We adjust the dimension of the final encoded feature output by adding a trainable linear layer behind the backbone. Once $G_1$ training is complete, we freeze the parameters in $G_1$ and utilize it for inference, subsequently training the parameters in $G_2$. The input of $G_2$ contains both the noised image $X_t$ at the resolution of this layer and the low-resolution image $\bar{Y}_1$ generated by $G_1$. We upsample $\bar{Y}_1$ to double its original resolution and then concatenate it with the noised image $X_t$ in the channel dimension before inputting it into $G_2$. Following a similar procedure, we train the generators $G_3$ and $G_4$ successively. Upon completion of training, we utilize all layers in the model for inference generation. The principle of our Diffusion version of HAIFIT can be referenced in Algorithm~\ref{alg_diff}.


\section{Experiments}
\label{sec:experiments}

\subsection{HAIFashion Dataset}
\label{sec:dataset}

\begin{figure}[t]
\centering
\includegraphics[width=0.96\columnwidth]{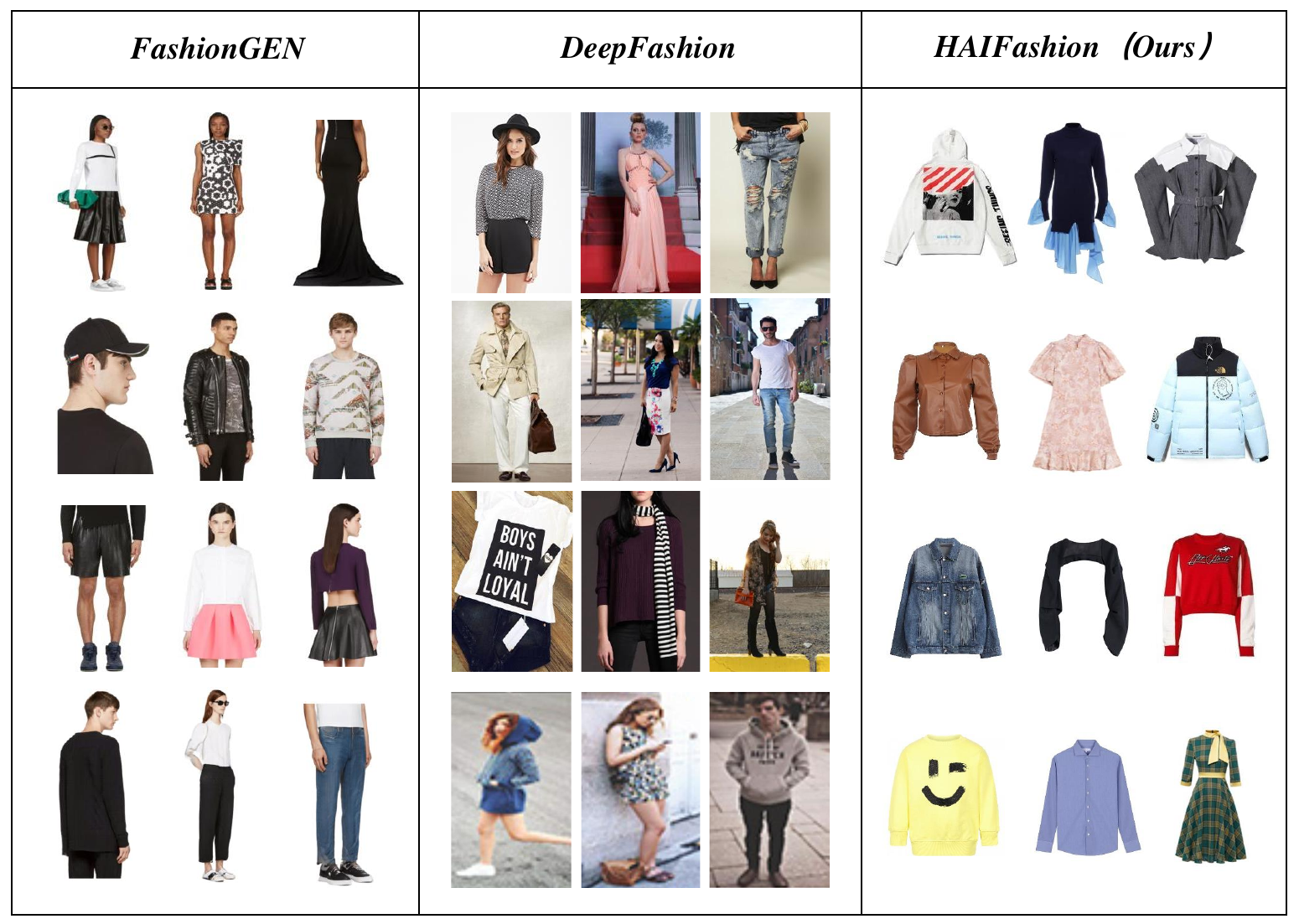}
\caption {Existing fashion datasets FashionGEN~\cite{rostamzadeh2018fashion} and DeepFashion~\cite{liuLQWTcvpr16DeepFashion} compared to our newly proposed dataset, HAIFashion.}
\label{fig:datset}
\end{figure}

Considering our objective of learning clothing sketch features and generating high-quality clothing images while preserving the designer's style and intricate details, it's imperative to construct a new clothing dataset comprising paired images and sketches for model training and testing. While directly hand-drawing a large dataset of garment image-sketch pairs by designers may seem challenging, a more feasible approach involves designers creating a small amount of clothing image-sketch paired data, which can then be expanded using an image-to-sketch generation algorithm. Thanks to the personalized sketch generation algorithm proposed by StyleMe~\cite{wu2023styleme}, this dataset construction becomes feasible. Thus, our dataset construction task focuses on building a fashion clothing image dataset for image-to-sketch generation.

Common publicly available fashion datasets like FashionGEN~\cite{rostamzadeh2018fashion} and DeepFashion~\cite{liuLQWTcvpr16DeepFashion}  predominantly feature realistic fashion images sourced directly from real-world photos. These images often include various backgrounds and individuals wearing the clothing in social settings. However, images with backgrounds pose significant challenges in generating single clothing sketches, which may not meet our requirements. The comparison between these datasets is illustrated in Fig.~\ref{fig:datset}. In this study, we introduce a self-collected clothing dataset named HAIFashion. HAIFashion comprises 3100 pairs of fashion clothing image-sketch datasets. We achieved this by crawling 3100 background-free fashion clothing images from the internet and employing the sketch generate algorithm~\cite{wu2023styleme} and the paired sketch-image dataset~\cite{jiang2024simple} to generate designer-style clothing sketches, facilitating our experiments.

\subsection{Implementation Details}
We implement our model on Pytorch and all experiments are performed on a single NVIDIA GeForce RTX4090 GPU. We utilize the Adam optimizer with a learning rate of 1e-4 for the generator and 5e-4 for the discriminator, the batch size is set to 8. The learning rates decayed at the halfway point every 100 epochs. The training process continued until the SSIM~\cite{wang2004image} showed no improvement for 10 iterations. For the dataset, we use our proposed HAIFashion (\S~\ref{sec:dataset}) and divide it into 2500 (600) for model training and testing respectively.

\begin{table*}[ht]
\caption{The Overall sketch-to-image generation performance on the HAIFashion dataset. ($\uparrow$ indicates the higher value, the better effect. $\downarrow$ indicates the lower value, the better effect.)}
\label{table:s2i}
\centering
\resizebox{\linewidth}{!}{
\begin{tabular}{lcccccccc} \toprule
\multirow{2}{*}{\textbf{Methods}} & \multirow{2}{*}{\textbf{Venue}} & \multirow{2}{*}{\textbf{Architecture}} & \multicolumn{3}{c}{\textbf{Structural}} & \multicolumn{3}{c}{\textbf{Perceptual}} \\ \cmidrule(lr){4-6} \cmidrule(lr){7-9}
& & & \quad \textbf{PSNR} $\uparrow$ \quad 
& \quad \textbf{SSIM} $\uparrow$ \quad 
& \quad \textbf{Overall} $\uparrow$ \quad 
& \quad \textbf{LPIPS} $\downarrow$ \quad 
& \quad \textbf{FID} $\downarrow$ \quad
& \quad \textbf{Overall} $\downarrow$ \quad \\ \midrule \midrule
BicycleGAN~\cite{zhu2017toward} & NeurIPS'17 & GAN-based
& 12.3482 & 0.4258 & 12.7740 & 0.1976 & 70.6381 & 70.8357 \\
pix2pix~\cite{isola2017image} & CVPR'17 & GAN-based
& \textbf{19.2761} & 0.3826 & \underline{19.6587} & 0.1180 & 48.3830 & 48.5010 \\
MUNIT~\cite{huang2018multimodal} & ECCV'18 & GAN-based
& 13.3265 & 0.4843 & 13.8108 & 0.1947 & 74.2733 & 74.4680 \\
UAGITI~\cite{alami2018unsupervised} & NeurIPS'18 & GAN-based
& 11.8259 & 0.0901 & 11.9160 & 0.2383 & 83.8309 & 84.0692 \\
pix2pixHD~\cite{wang2018high} & CVPR'18 & GAN-based
& 16.1863 & \underline{0.5622} & 16.7485 & \underline{0.1026} & \underline{48.0041} & \underline{48.1067} \\
USTP~\cite{liu2020unsupervised} & ECCV'20 & GAN-based
& 13.2456 & 0.5126 & 13.7582 & 0.1429 & 58.8756 & 59.0185 \\
UGATIT~\cite{kim2020u-gat-it} & ICLR'20 & GAN-based
& 12.8841 & 0.4799 & 13.3640 & 0.1783 & 83.9190 & 84.0973 \\ 
IrwGAN~\cite{xie2021unaligned} & ICCV'21 & GAN-based
& 17.2878 & 0.4555 & 17.7433 & 0.1061 & 71.7609 & 71.8670 \\
Palette~\cite{saharia2022palette} & SIGGRAPH'22 & DM-based
& 7.8216 & 0.0787 & 7.9003 & 0.3591 & 116.9303 & 117.2894 \\
BBDM~\cite{li2023bbdm} & CVPR'23 & DM-based
& 14.9788 & 0.5343 & 15.5131 & 0.1676 & 59.7919 & 59.9595 \\ \midrule
\textbf{HAIFIT} & Ours & GAN-based
& \underline{19.0660} & \textbf{0.6338} & \textbf{19.6998} & \textbf{0.0778} & \textbf{28.5032} & \textbf{28.5810} \\
\bottomrule
\end{tabular}
}
\end{table*}

\subsection{Evaluation Metrics}
We employ four commonly used quantitative evaluation metrics:
\begin{itemize}
\item \textbf{PSNR} (Peak Signal-to-Noise Ratio) assesses image quality by comparing the peak signal-to-noise ratio between the original image and the generated image. Higher PSNR values indicate smaller differences between the images and higher-quality generated images.
\item \textbf{SSIM} (Structural Similarity Index)~\cite{wang2004image} evaluates image quality by comparing the structural similarity between the original image and the generated image. It considers similarities in brightness, contrast, and structure to calculate the final score. Larger SSIM values signify higher similarity between the images and better generation effects.
\item \textbf{LPIPS} (Learned Perceptual Image Patch Similarity)~\cite{zhang2018unreasonable} assesses image quality by utilizing a deep learning model and the L2 norm to calculate the perceptual difference between the generated image and the original image in the feature space. Lower LPIPS values suggest better-quality images.
\item \textbf{FID} (Fréchet Inception Distance)~\cite{heusel2017gans} evaluates image quality by comparing the statistical distribution in the feature space of the generated image with that of the original image. It employs the Fréchet distance to measure the similarity between the distributions. Lower FID values indicate higher-quality resulting images.
\end{itemize}
The first two metrics primarily gauge the completeness and structural similarity of the generated images, whereas the latter two metrics focus more on assessing the diversity, authenticity, and perceptual differences of the generated images.

\subsection{Baselines}
In our experiments, we compared our method with ten state-of-the-art baseline approaches. All models undergo identical preprocessing steps and are trained directly following the official open-source code to ensure a fair comparison. 
\begin{itemize}
\item \textbf{BicycleGAN}~\cite{zhu2017toward} focuses on modeling a distribution of possible outputs in a conditional generative model, enhancing the diversity of generated images.
\item \textbf{pix2pix} \cite{isola2017image} implements image translation using conditional GAN to convert input images between different domains.
\item \textbf{MUNIT} \cite{huang2018multimodal} achieves image synthesis by decomposing image representations into content and style properties.
\item \textbf{UAGITI} \cite{alami2018unsupervised} incorporates attention networks to minimize divergence between relevant parts of the data-generating distribution.
\item \textbf{pix2pixHD} \cite{wang2018high} utilizes a novel adversarial loss and new multi-scale generator and discriminator architectures in conditional GAN for high-resolution image synthesis.
\item \textbf{USTP} \cite{liu2020unsupervised} learns from unpaired sketch and photo data and tackles synthesis through shape translation and content enrichment stages.
\item \textbf{UGATIT} \cite{kim2020u-gat-it} employs an attention module and a learnable normalization function to enhance the quality and diversity.
\item \textbf{IrwGAN}~\cite{xie2021unaligned} introduces the concept of importance-based reweighting to select images and devises a method to learn the weights autonomously while simultaneously conducting image translation.
\item \textbf{Palette} \cite{saharia2022palette} proposes a unified framework for image-to-image translation based on conditional diffusion models, demonstrating the importance of self-attention.
\item \textbf{BBDM} \cite{li2023bbdm} presents a novel image-to-image translation approach utilizing the Brownian bridge diffusion model, which enables direct learning of domain translation through a bidirectional diffusion process.
\end{itemize}

\subsection{Comparison Results}

\begin{figure*}[t]
\centering
\includegraphics[width=0.98\linewidth]{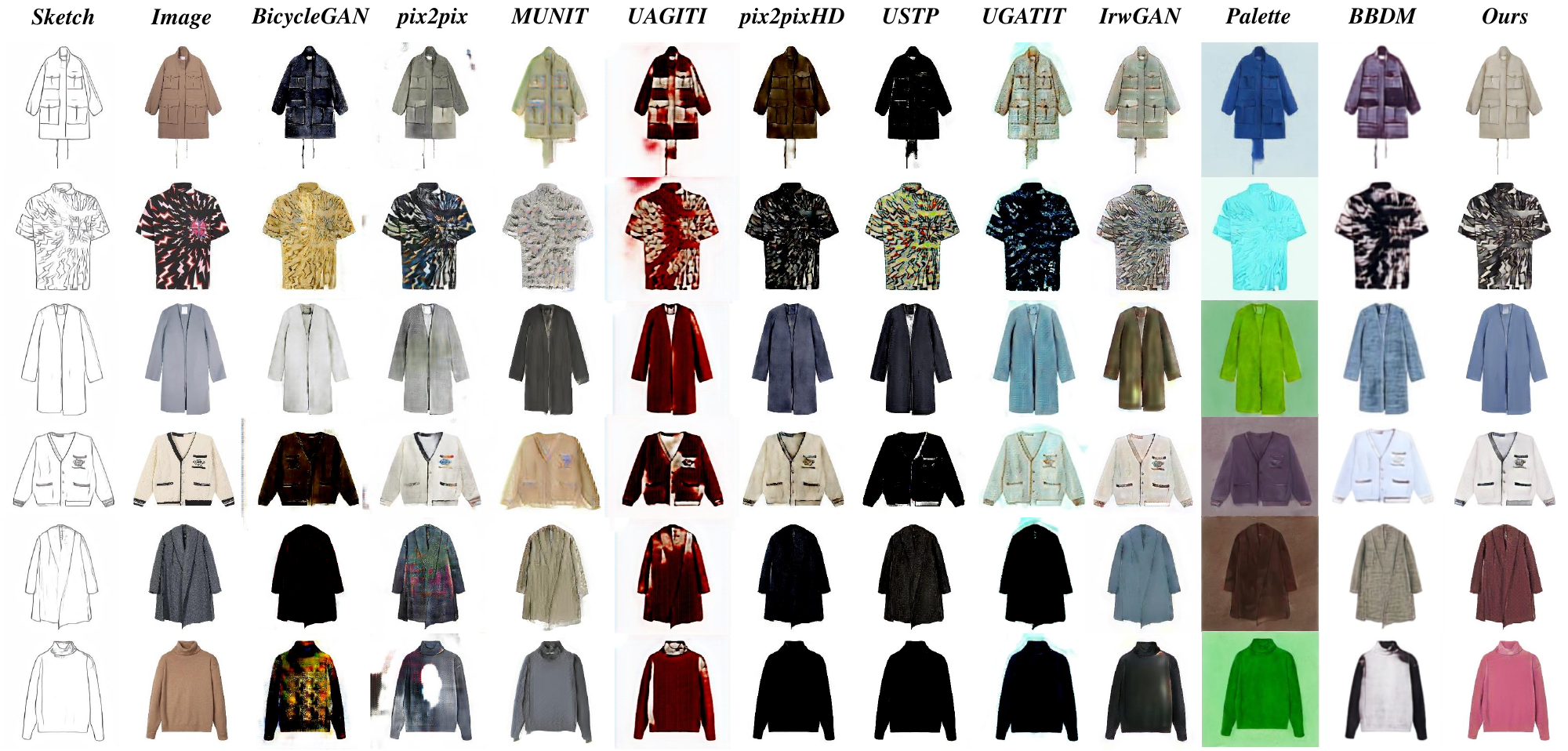}
\caption {The comparison of sketch-to-image generation from the baselines and our method. The first column is the input sketch, the second column is the corresponding reference image, and the other columns are images generated by different methods.}
\label{fig:s2i_gen}
\end{figure*}

\textbf{Quantitative Experiments. }
We conducted quantitative evaluations to compare both the realism and faithfulness of the generated images. The quantitative results are summarized in Table~\ref{table:s2i}, despite achieving the second position in the PSNR index, it is important to note that this result is primarily due to the nature of PSNR, which emphasizes the signal-to-noise ratio and is sensitive to local noise and distortion in the image. HAIFIT, on the other hand, prioritizes retaining more details from the input sketches, including the initial outline, which may lead to a slight decrease in the PSNR index. However, compared to other benchmarks, HAIFIT all achieved first place. In the SSIM metric, our method outperforms the second-placed pix2pixHD model by nearly 13\%. For metrics that are more aligned with human visual perception, our method enhances the LPIPS and FID indicators by about 24\% and 41\% respectively compared to the second-place model, demonstrating superior performance overall. In summary, HAIFIT exhibits excellent performance across all four metrics. This demonstrates the effectiveness of our method in sketch-to-image generation. This success can be attributed to the integration of our proposed MFFE and pyramid generator with CSCM. The MFFE helps our model learn object sketch contours and consider correlations between adjacent lines from different directions, enhancing stroke variation perception. Meanwhile, the pyramid generator with CSCM can capture specific details in key regions of complex sketches, facilitating the learning of the complete structure and fine-tuning of intricate details. The results confirm that our model can well adapt to sketches with specific detailed styles and generate more realistic images.

\noindent \textbf{Qualitative Experiments. }
As shown in Fig.~\ref{fig:s2i_gen}, it is evident that our proposed method outperforms other baselines by preserving sketch details and delivering high-definition diverse images. For example, while BcycleGAN, USTP, pix2pixHD, and UGATIT occasionally generate satisfactory results, such as the second row in cycleGAN and USTP, and the third row in pix2pixHD and UGATIT, most of their generated images tend to exhibit a single black tone and lose much of the sketch's stroke details. Although pix2pix, MUNIT, and IrwGAN methods show better generation effects with good color diversity, they still face numerous challenges. For instance, pix2pix generates colors outside the designated area in the first and sixth rows, and displays abnormal color distribution in the fifth row. MUNIT often loses significant sketch details, causing the edges of the generated images to appear overly smooth and blurry. Moreover, IrwGAN suffers from uneven color distribution in the first, second, and third rows, and a loss of sketch details in the fifth and sixth rows. The UGAIT method consistently shows a red tone in its generated images, along with issues like uneven color distribution, coloring outside the designated areas, and blurred details. In addition, when comparing diffusion models, we find that Palette has serious background coloring issues, making the image background color match the clothing color, and also faces problems with uneven color distribution and loss of sketch details. BBDM, while exhibiting uniform color distribution within contours and avoiding coloring outside the areas, still struggles with retaining fine details in the generated images. Overall, existing methods still struggle with capturing specific complex details in sketches and generating high-definition, diverse images. In contrast, our proposed method excels in handling complex sketches, producing images with more realistic visual effects.

\begin{figure}[t]
\centering
\includegraphics[width=0.8\columnwidth]{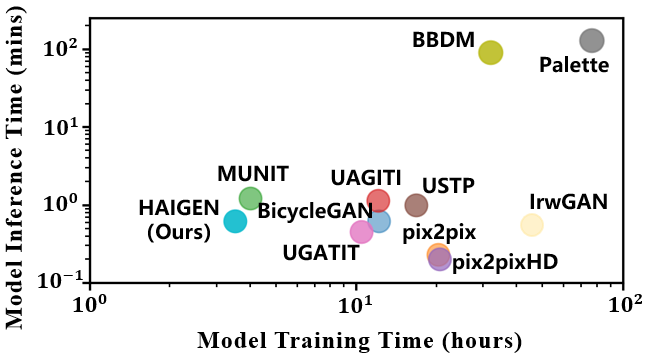}
\caption{Comparison of model training and inference times. The coordinate axes adopt a logarithmic value method to better show the positional relationship between them.}
\label{fig:time}
\end{figure}

\noindent \textbf{Time Costs. }
Fig.~\ref{fig:time} illustrates the model training and inference times for our model and the ten baselines. Our model exhibits a clear advantage in training time compared to GAN-based methods. Additionally, when compared to the methods based on the Diffusion model, our model demonstrates an inference time improvement of over 100 times. This significant time-saving advantage can greatly benefit designers by reducing costs, improving work efficiency, and increasing productivity.

\subsection{Ablation study}
We conduct ablation studies to analyze the effectiveness of (i) the Abstract Feature Representation Module (AFRM) and (ii) the Cross-level Skip Connection Module (CSCM).

\noindent \textbf{Quantitative Experiments. }
As indicated in Table~\ref{tab:s2iab}, the removal of the CSCM between different layers of HAIFIT negatively impacts our model's performance. CSCM is essential in considering feature correlations across scales and combining different receptive fields, which helps preserve important semantic details. The increase in the PSNR index after removing CSCM is attributed to the loss of sketch detail features, resulting in smoother perceived details. Moreover, removing the AFRM also led to a decline in our model's performance. AFRM learns long-term dependencies from different perspectives of the sketch using the LSTM module. It takes into account the correlation between adjacent lines in the sketch to increase the details and ensure the consistency of the generated image. Notably, when replacing the LSTM module with the Transformer~\cite{vaswani2017attention} module, all four performance indicators of the model deteriorated. This decline can be attributed to the temporal dependence offered by the LSTM module, ensuring consistent feature extraction across all directions of the sketch and yielding a more intricate and coherent feature representation. In contrast, the Transformer module, relying on the self-attention mechanism, captures relationships within sequences but lacks strong modeling capabilities in preserving the sequence's order of features. This outcome underscores the significance of considering the temporal correlation between lines to enhance model performance. Based on the above analysis, the AFRM and CSCM modules enhance our model's capability to handle complicated sketches. Consequently, HAIFIT enables the generation of high-quality and diverse clothing images while maintaining the complex details of the sketches.

\begin{table}[t]
\caption{The quantitative experimental results on ablation study.}
\label{tab:s2iab}
\centering
\resizebox{\columnwidth}{!}{
\begin{tabular}{ccccc} \toprule
\multirow{2}{*}{\textbf{Methods}} & \multicolumn{2}{c}{\textbf{Structural}} & \multicolumn{2}{c}{\textbf{Perceptual}} \\ \cmidrule(lr){2-3} \cmidrule(lr){4-5}
&\textbf{PSNR} $\uparrow$ &\textbf{SSIM} $\uparrow$ &\textbf{LPIPS} $\downarrow$ &\textbf{FID} $\downarrow$ \\ \midrule \midrule
w/o AFRM and CSCM & \underline{19.7086} & 0.6324 & 0.0812 & 30.5260 \\ 
w/o AFRM  & 19.0474 & 0.6281 & 0.0821 & 31.8295 \\
w/o CSCM  & \textbf{19.7189} & 0.6294 & 0.0802 & 29.9711 \\
Transformer~\cite{vaswani2017attention}  & 18.7738 & 0.6259 & 0.0814 & 31.1292 \\ 
\textbf{HAIFIT (full)}  & \underline{19.0660} & \textbf{0.6338} & \textbf{0.0778} & \textbf{28.5032} \\ \bottomrule
\end{tabular}
}
\end{table}

\begin{figure}[t]
\centering
\includegraphics[width=0.95\columnwidth]{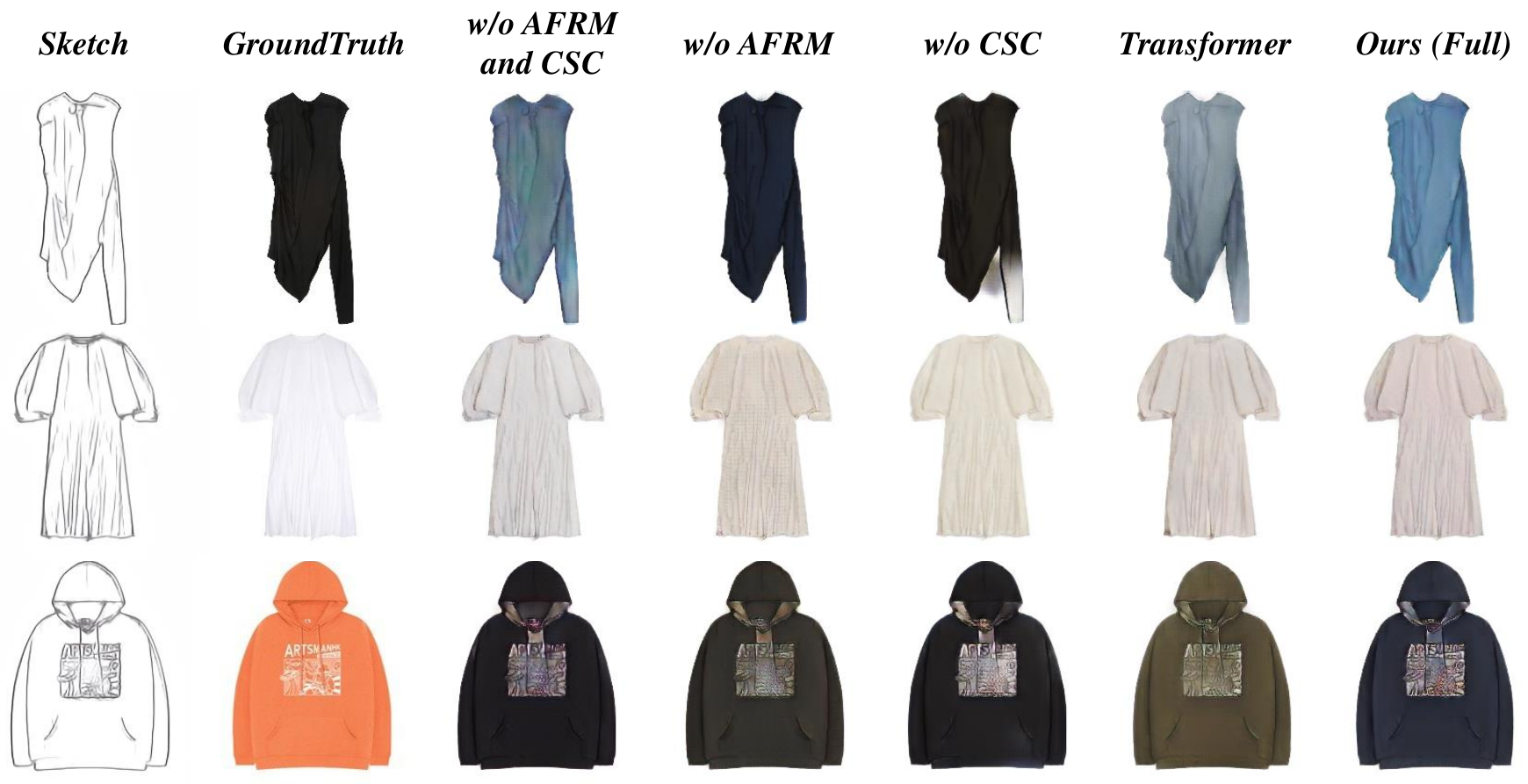}
\caption{Qualitative results of the ablation study on HAIFIT.}
\label{fig:s2iab}
\end{figure}

\noindent \textbf{Qualitative Experiments. }
The qualitative results of the ablation study on sketch-to-image generation are presented in Fig.~\ref{fig:s2iab}. It can be observed that removing the AFRM results in the loss of some details, causing the garment patterns to become blurred in the third row. When the CSCM module is removed, the images in the first row exhibit issues such as coloring outside the edges and a loss of sketch stroke details. These problems become more pronounced when both modules are removed simultaneously, leading to noticeable unevenness in color distribution in some images. In contrast, replacing the LSTM module with the Transformer module yielded better visual effects than other situations, such as a gradient effect on the clothing in the first row. However, there are still challenges in retaining sketch details. These findings collectively confirm the effectiveness of each module in our model.


\section{Conclusions}
\label{sec:conclusion}

In this paper, we introduce HAIFIT, a creative sketch-to-image generation model tailored for fashion design. We effectively enhance both the global consistency of sketch feature vectors and the richness of intricate textures in the latent space through the proposed MFFN and CSCM. This enhancement enables the generation of high-quality, realistic clothing images. Furthermore, We present a fashion clothing dataset HAIFashion for image translation to fill the gap in this field. We conduct extensive experiments comparing our method with state-of-the-art approaches. The qualitative and quantitative results robustly demonstrate the effectiveness of our proposed methodology in preserving sketch details and generating realistic images.

\bibliographystyle{plain}
\bibliography{main}

\end{document}